\documentclass[letterpaper, 10 pt, conference]{ieeeconf}  

\IEEEoverridecommandlockouts                              

\usepackage{graphicx} 
\usepackage{amsmath} 

\title{\LARGE \bf
UF-RNN: Real-Time Adaptive Motion Generation Using Uncertainty-Driven Foresight Prediction
}

\author{Hyogo Hiruma$^{1, 2}$, Hiroshi Ito$^{1, 2}$, and Tetsuya Ogata$^{1, 3}$
\thanks{*This work was supported by JST [Moonshot R\&D][Grant Number JPMJMS2031].}
\thanks{$^{1}$Hyogo Hiruma, Hiroshi Ito and Tetsuya Ogata are with the Department of
              Intermedia Art and Science, Waseda University, Tokyo, Japan
        hiruma@idr.ias.sci.waseda.ac.jp, hiroshi.ito.ws@hitachi.com, ogata@waseda.jp,
        $^{2}$Hyogo Hiruma and Hiroshi Ito are with Research Development Group, Hitachi, Ltd. Tokyo, Japan,
        $^{3}$Tetsuya Ogata is with the National Institute of Advanced Industrial Science and Technology (AIST), Tokyo, Japan}
}

\begin{document}

\maketitle
\thispagestyle{empty}
\pagestyle{empty}

\begin{abstract}

Training robots to operate effectively in environments with uncertain states—such as ambiguous object properties or unpredictable interactions—remains a longstanding challenge in robotics. Imitation learning methods typically rely on successful examples and often neglect failure scenarios where uncertainty is most pronounced. To address this limitation, we propose the Uncertainty-driven Foresight Recurrent Neural Network (UF-RNN), a model that combines standard time-series prediction with an active “Foresight” module. This module performs internal simulations of multiple future trajectories and refines the hidden state to minimize predicted variance, enabling the model to selectively explore actions under high uncertainty. We evaluate UF-RNN on a door-opening task in both simulation and a real-robot setting, demonstrating that, despite the absence of explicit failure demonstrations, the model exhibits robust adaptation by leveraging self-induced chaotic dynamics in its latent space. When guided by the Foresight module, these chaotic properties stimulate exploratory behaviors precisely when the environment is ambiguous, yielding improved success rates compared to conventional stochastic RNN baselines. These findings suggest that integrating uncertainty-driven foresight into imitation learning pipelines can significantly enhance a robot’s ability to handle unpredictable real-world conditions.
 
\end{abstract}

\section{INTRODUCTION}

Robots deployed in real-world environments must be capable of adapting to diverse and dynamic situations. Traditional rule-based approaches have been widely used to program robotic behaviors, but they often lack the flexibility to generalize across varying conditions \cite{schaal1999imitation, osa2018algorithmic}. With the advent of deep learning, data-driven methods have emerged as a promising alternative, enabling robots to adapt to novel situations with minimal manual engineering \cite{levine2016end}. In particular, imitation learning has gained significant attention as a technique that allows robots to acquire task-specific behaviors from a limited number of demonstrations \cite{ito2022efficient}.

A key challenge in achieving adaptability is handling situations with high uncertainty. Uncertainty arises when the appropriate action cannot be uniquely determined given the current state. For example, when opening a door, the correct action depends on the opening direction of the door, which may not be immediately known \cite{zhang2024learning}. To address this challenge, an agent must (1) recognize that the environment is uncertain and (2) actively take actions to reduce this uncertainty and determine the correct course of action.

Existing approaches attempt to address these challenges through various mechanisms. Action Chunking Transformer (ACT) \cite{zhao2023learning} and Diffusion Policy models \cite{chi2023diffusion}, for example, leverage large datasets to learn multi-modal transition dynamics, allowing adaptive behavior across different scenarios. Diffusion Policy, in particular, formulates policy generation as a denoising process, exploring multiple potential actions before selecting the most suitable one. However, such methods often lack an explicit mechanism for directing exploration toward uncertainty reduction, leading to suboptimal information gathering and inefficient policy learning.

\begin{figure}[t]
  \centering
  \includegraphics[width=0.9\linewidth]{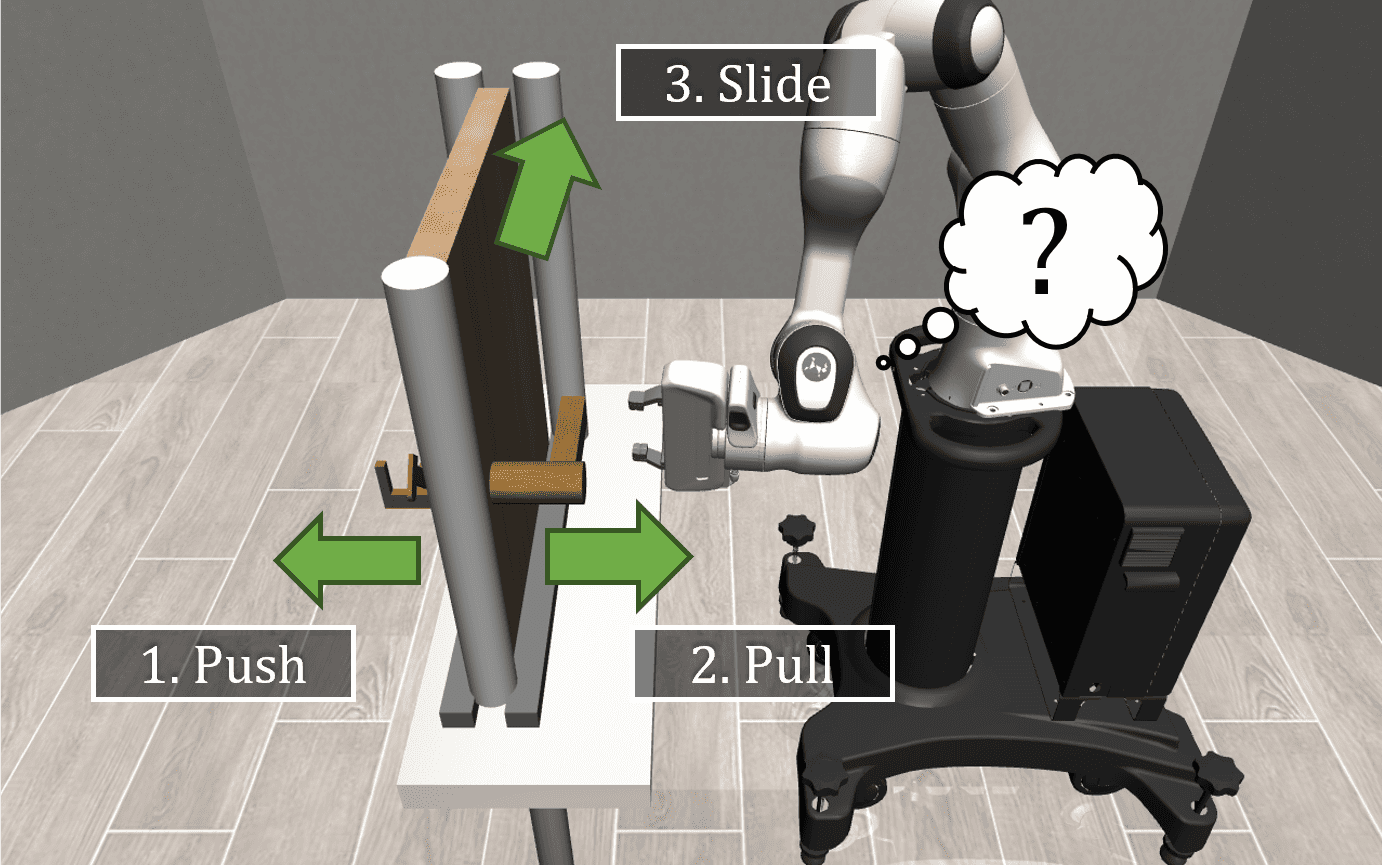}
  \caption{Example of a task that embrace high uncertainty. Opening doors
  may require the robot to interact before understanding which way the door can 
  be opened, especially when it is not visually distinguishable. }
  \label{fig:sim_env}
\end{figure}

To overcome these limitations, we draw inspiration from the concept of Active Inference, as formulated by Friston et al. \cite{friston2017active}. Active Inference posits that biological agents minimize Expected Free Energy (EFE) \cite{friston2006free} to optimize their actions. EFE consists of two key components: (1) the maximization of immediate rewards (exploitation) and (2) the acquisition of new information to reduce future uncertainty (exploration). In uncertain environments, human decision-making tends to prioritize the latter, favoring actions that yield informative outcomes. This principle aligns with the well-known exploration-exploitation dilemma in reinforcement learning \cite{Auer2002Finite-time}.

\begin{figure*}[t]
    \centering
    \includegraphics[width=\linewidth]{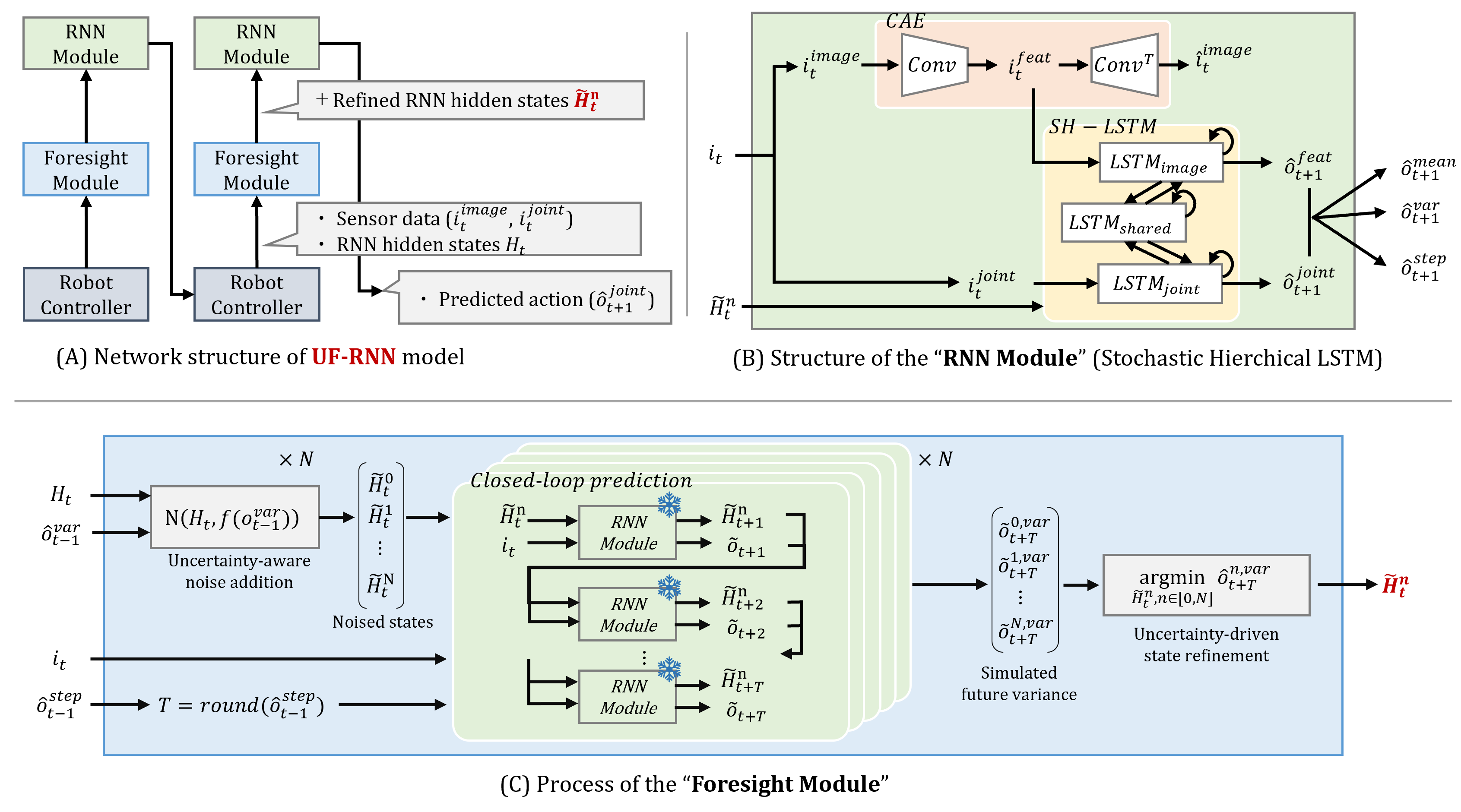}
    \caption{Network structure of the proposed \textit{UF-RNN} model. (A) The model consists of an RNN module and a novel Foresight Module, which is used to refine the RNN hidden states through internal simulations. (B) The structure of RNN module, which uses a stochastic hierarchical RNN (SH-RNN) for predicting the expected sensor
    values at the next time step. (C) describes the process of Foresight Module which re-directs the RNN hidden states to guide to less uncertain states via internal simulation (closed-loop prediction).  }
    \label{fig:model_structure}
\end{figure*}

Building on this foundation, we propose the \textbf{Uncertainty-Driven Foresight Recurrent Neural Network (UF-RNN)}, a model that enables robotic agents to actively reduce uncertainty through future-aware decision-making. Specifically, UF-RNN predicts multiple possible future scenarios before executing an action and selects the policy that minimizes future uncertainty. Unlike traditional approaches, which rely on predefined state transition models or environment constraints, UF-RNN learns these dynamics through its own training process, making it applicable to continuous and unstructured environments.

The key contributions of our work are as follows:
\begin{enumerate}
\item We introduce a novel uncertainty-aware policy selection mechanism based on Expected Free Energy minimization.
\item We propose UF-RNN, a model that forecasts multiple future scenarios and selects actions that reduce long-term uncertainty.
\item We demonstrate the applicability of UF-RNN to real-world robotic tasks, showing its ability to generalize beyond predefined state-action mappings and to unforeseen disturbances.
\end{enumerate}

Through extensive evaluations, we validate that UF-RNN improves exploration efficiency and policy robustness in uncertain environments. Our approach opens new avenues for integrating cognitive principles into robotic decision-making, bringing us closer to human-like adaptability in real-world settings.

\section{Related Research}

\subsection{Uncertainty Estimation}

Uncertainty estimation is vital for robust robotic decision-making, as it helps agents to handle incomplete or noisy observations. Classical approaches, such as Kalman Filters (KF) and Particle Filters (PF), have long been used for self-localization and motion planning \cite{fox2000probabilistic, Rigatos2010Extended}. Although these methods excel at modeling well-defined system dynamics, they often depend on hand-crafted assumptions that do not readily generalize to unstructured environments.

Learning-based methods have emerged to address these limitations by allowing models to learn distributions directly from data. For instance, Bayesian Neural Networks (BNNs) capture epistemic uncertainty through distributions over weights \cite{blundell2015weight}, although their scalability can be challenging. Deep Ensembles \cite{lakshminarayanan2017simple} and Disagreement Models \cite{pathak2019self} approximate epistemic uncertainty by training multiple models with varying priors, making them suitable for domains where unseen scenarios are frequent. Meanwhile, Diffusion Policy \cite{chi2023diffusion} uses a denoising diffusion model to capture aleatoric uncertainty, offering stochastic action sampling in imitation learning tasks. However, purely diffusion-based approaches may overlook epistemic uncertainty \cite{Chan2024EstimatingEA}, which is crucial for environments that differ from training data.

In parallel, recurrent architectures have been proposed to handle temporal uncertainties. Stochastic RNNs (S-RNNs) \cite{murata2013learning,chung2015recurrent,ahmadi2019novel} predict both the mean and variance of future states, enabling robots to anticipate multiple possible outcomes in dynamic settings. These methods facilitate the modeling of both deterministic dynamics and random fluctuations in sensory inputs, yet they do not always provide a single framework that combines aleatoric and epistemic components under data-scarce conditions.

\subsection{Uncertainty-Guided Inference}

Despite progress in uncertainty modeling, translating these estimates into effective policies remains a challenge. In reinforcement learning (RL), curiosity-driven methods \cite{pathak2017curiosity} reward agents to predict environmental dynamics, thus encouraging exploration in areas of high epistemic uncertainty. Bayesian RL \cite{ghavamzadeh2015bayesian} offers a theoretical foundation for maintaining posterior distributions over rewards and transitions, although real-world applications are often restricted to low-dimensional tasks.

Active Inference \cite{friston2017active,tschantz2020reinforcement} provides another perspective by casting decision-making as a variational free-energy minimization problem, allowing agents to reduce uncertainty about future states. However, the most successful applications of Active Inference in robotics focus on tasks with comparatively limited action spaces \cite{ueltzhoffer2018deep,schwartenbeck2019computational}. Scaling it to high-dimensional or continuous domains \cite{oliver2021empirical}, and integrating it with imitation learning remain open challenges.

In summary, existing work underscores the importance of robust uncertainty estimation and its potential to guide exploration and decision-making. However, a gap persists in building unified approaches that handle both aleatoric and epistemic uncertainty within data-scarce imitation learning contexts. Addressing this gap is critical for enabling reliable and uncertainty-aware robot behavior in complex real-world scenarios.

\section{Method}
\label{sec:method}

\subsection{Overall Model Description}
The proposed UF-RNN model follows the prediction scheme of deep predictive learning\cite{suzuki2023deep}, where the RNN modules learn the sensorimotor dynamics of the robot and generate the robot actions based on observed sensor data (Fig.~\ref{fig:model_structure}(A)). The input $i_t$ is the sensor data observed in the current timestep $t$, and the output $o_{t+1}$ and $\widehat{o}_{t+1}$ are the ground truth and predicted sensor data in the next timestep, respectively. The motion is generated by continuously predicting the next sensor values and applying them to the robot’s controller. The sensor data $i_t$ may include multiple modalities of robot data, where in this implementation, joint angles $i_t^{joint}$ and visual feature vectors $i_t^{image}$ are fed into the RNN. The proposed model also adds a \textit{foresight module}, which refines the model’s hidden states $H_t$ to $\widehat{H}_t$ such that the future variance, or uncertainty, across all modalities is minimized.

\subsection{RNN Module}
Fig.~\ref{fig:model_structure}(B) shows the details of the RNN module. It consists of a Convolutional Autoencoder (CAE) and a Stochastic Hierarchical Long Short-Term Memory (SH-LSTM). A pre-trained CAE is used to compress high-dimensional visual inputs $i_t^{image}$ into low-dimensional representations $i_t^{feat}$, while the SH-LSTM predicts time-series sensor data.

\begin{figure*}[tbp]
    \centering
    \includegraphics[width=\linewidth]{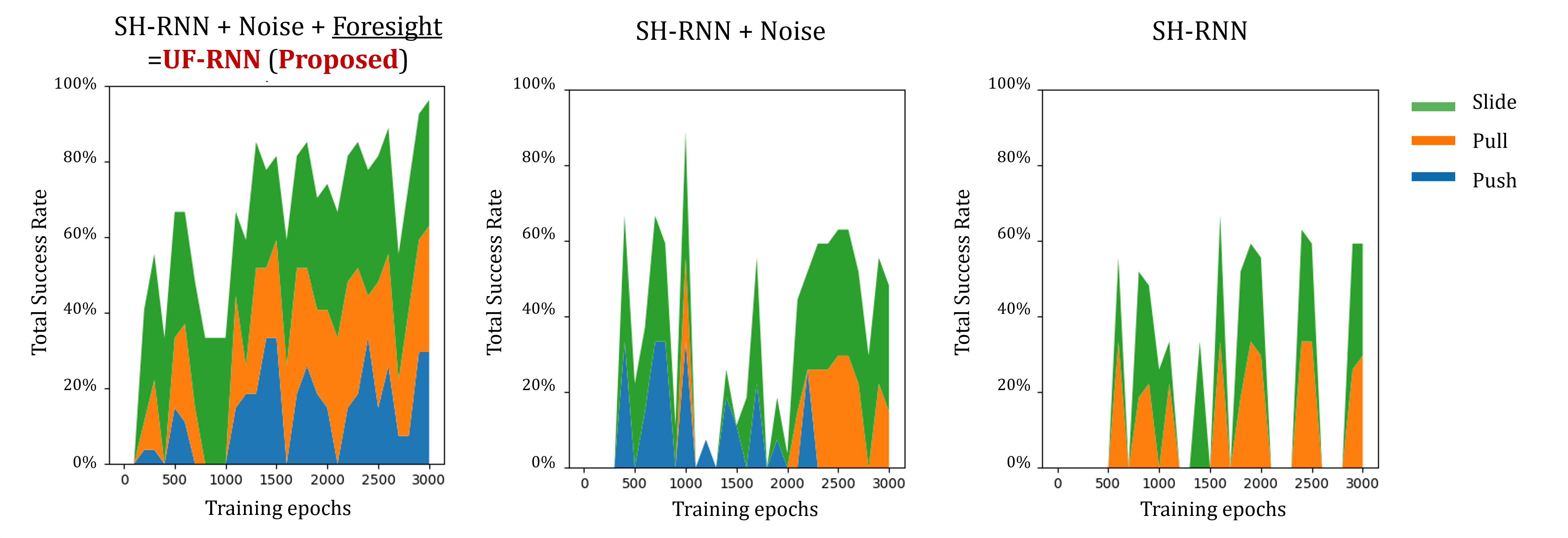}
    \caption{Comparison of success rates per training model. The graph shows the transition of success rates using the trained model every 100 epochs, up to 3000. The graph is a stacked plot color coded by door types, showing the success rates of each type per 10 trials.}
    \label{fig:sim_success_rate}
\end{figure*}

\begin{figure*}[tbp]
    \centering
    \includegraphics[width=\linewidth]{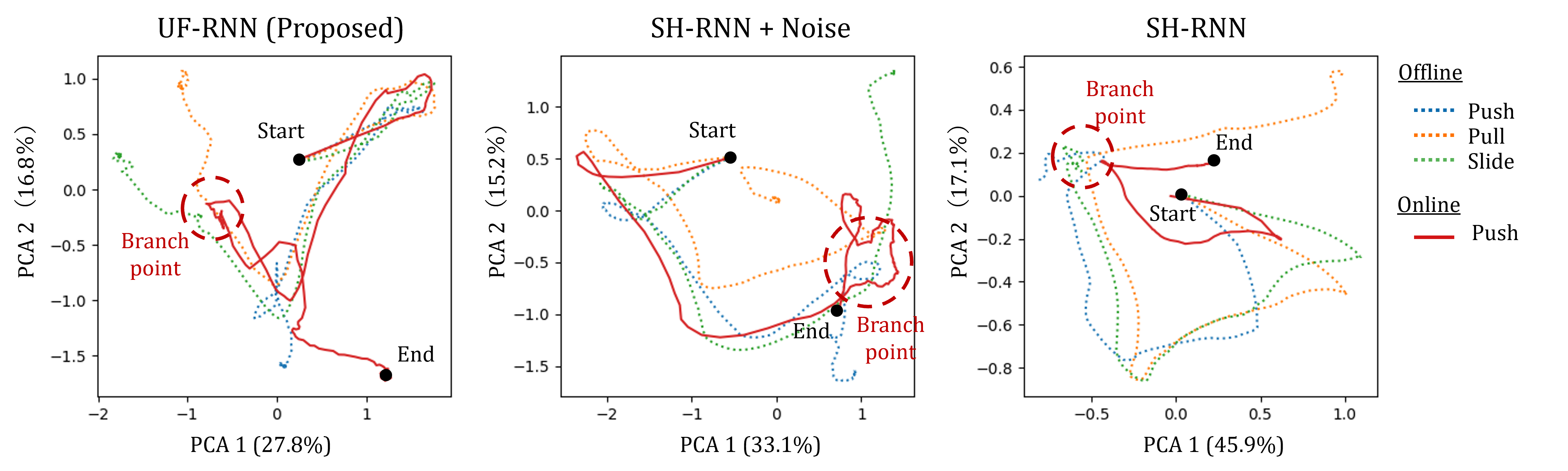}
    \caption{Comparison on transitions of RNN hidden states per training model type. The graph plots the hidden states of $LSTM_{shared}$, which is compressed to two dimensions using principal component analysis. The dotted lines show prediction of three motions on pre-collected data (offline), and the solid line shows the prediction on pushing motion during inference (online). }
    \label{fig:sim_pca}
\end{figure*}

Here, SH-LSTM is a combined architecture that incorporates (a) stochastic predictive properties from S-RNN~\cite{murata2013learning}, (b) a hierarchical structure inspired by Multiple timescale RNN (MT-RNN)~\cite{yamashita2008emergence} to capture multi-timescale features, and (c) partitioning and unifying structure to capture intra- and intermodal dynamics of RNN modules from Active Attention RNN ($A^2 RNN$)~\cite{hiruma2022deep}. Specifically, the lower-layer LSTMs ($LSTM_{image}$ and $LSTM_{joint}$) handle raw sensor data dynamics (short timescale), while the higher-layer LSTM ($LSTM_{shared}$) manage more abstract, long-range information. By training the model in this hierarchical stochastic setting, the lower layer captures fast sensor dynamics, and the higher layer captures extended temporal dependencies.

Each lower layer LSTM predicts three quantities:
\begin{enumerate}
  \item The mean of the next sensor data $\widehat{o}_{t+1}^{mean}$
  \item The variance of the next sensor data $\widehat{o}_{t+1}^{var}$
  \item A scalar value $T = \widehat{o}_{t+1}^{step}$, which determines how many future steps are simulated in the foresight module. This $T$ can differ between sensor modalities, and we take the maximum across modalities as the effective horizon.
\end{enumerate}

\subsection{Foresight Module}
\label{sec:fs_module}
Fig.~\ref{fig:model_structure}(C) shows the detailed structure of the foresight module. This module generates \textit{foresight}, that is, the expected future states $(o_{t+T}^{mean},\, o_{t+T}^{var})$ up to $T$ timesteps ahead, by performing \textit{closed-loop prediction}. In closed-loop prediction, the model recursively feeds its own predicted outputs back as input for the next timestep, allowing it to internally simulate how the environment might proceed from its current hidden state $H_t$. The simulation is based on the dynamics that are captured within the RNN modules, as well as the model's belief, which is the RNN's temporal context information that is built along its continuous predictions. This process can be executed at any point in time, both during training (to improve the learned dynamics) and at run-time (for decision making).

The following describes the key modules to enable uncertainty-aware internal exploration in foresight prediction.

\subsubsection{Multiple Noise Perturbations}
To explore different future trajectories, the module applies $n$ distinct Gaussian noise perturbations to the hidden state $H_t$. Concretely:
\begin{enumerate}[label=\arabic*.]
  \item Sample $N$ perturbed hidden states $\widetilde{H}_t^n$ from $\mathcal{N}\bigl(H_t^n,\,f(o_{t-1}^{var})\bigr)$, where $f(\cdot)$ normalizes the variance of the previous timestep into a range $[0.05,\,0.15]$.
  \item Perform closed-loop predictions from each $\widetilde{H}_t^n$ for $T$ steps, yielding $N$ predicted future states $(o_{t+T}^{n,mean},\,o_{t+T}^{n,var})$.
  \item Select the hidden state $\widetilde{H}_t$ that lowered predicted variance across modalities
\end{enumerate}

The hidden state is selected based on how large the variance decreased after $T$ steps of foresight:
\[
    \widetilde{H}_t = \underset{\widetilde{H}_t^n} {\operatorname{argmax}}
    \,(o_{t}^{n,var} - o_{t+T}^{n,var}),
\]
where $o_{t+T}^{n,var}$ represents the variance in timestep $t+T$ from the $n$-th perturbation. This approach is inspired by the Expected Free Energy (EFE) principle in the Free-energy framework~\cite{friston2006free}, where agents prioritize actions that reduce future uncertainty.

\subsubsection{Adaptive Noise Intensity}
During foresight prediction, the intensity of the random noise is proportional to the previously predicted variance $\widehat{o}_{t-1}^{var}$. Hence, when the model detects high uncertainty, it increases noise intensity to explore more diverse future trajectories, while low uncertainty results in reduced noise for more conservative predictions. By adjusting noise according to the variance, the system aims to balance exploration and exploitation in an uncertainty-aware manner.

\subsection{Training Procedure}
\label{sec:loss}
We train the UF-RNN in an end to end format using visual images and joint angle data as input modalities. The overall loss function follows \cite{murata2013learning} which incorporates two main terms: (1) an reconstruction loss with predicted variance, and (2) a stochastic regularization term. Formally, we define the total loss as

\[
\mathcal{L} = \sum_{t=0}^{T} \Biggl(
  - \frac{\bigl(o_t^{\text{mean}} - \hat{o}_t^{\text{mean}}\bigr)^2}{2\,o_t^{\text{var}}}
  \;-\;
  \frac{\ln\!\bigl(2\pi\,\hat{o}_t^{\text{var}}\bigr)}{2}
\Biggr).
\]

where the weights are trained using full back propagation through time. The reconstruction loss measures the deviation between the predicted future sensor states and those observed during the expert trajectories. The prediction errors are divided by the predicted variance of each time step, to attenuate the effect of data that have high uncertainty. The stochastic regularization term constrains the predicted variance representations to remain stable and consistent over time. By combining these two terms, the UF-RNN model jointly learns to reproduce expert behaviors and maintain coherent probabilistic structures in its internal dynamics, leading to robust uncertainty-aware predictions. 
\section{Simulator Experiment}

\begin{figure*}[tbp]
    \centering
    \includegraphics[width=0.8\linewidth]{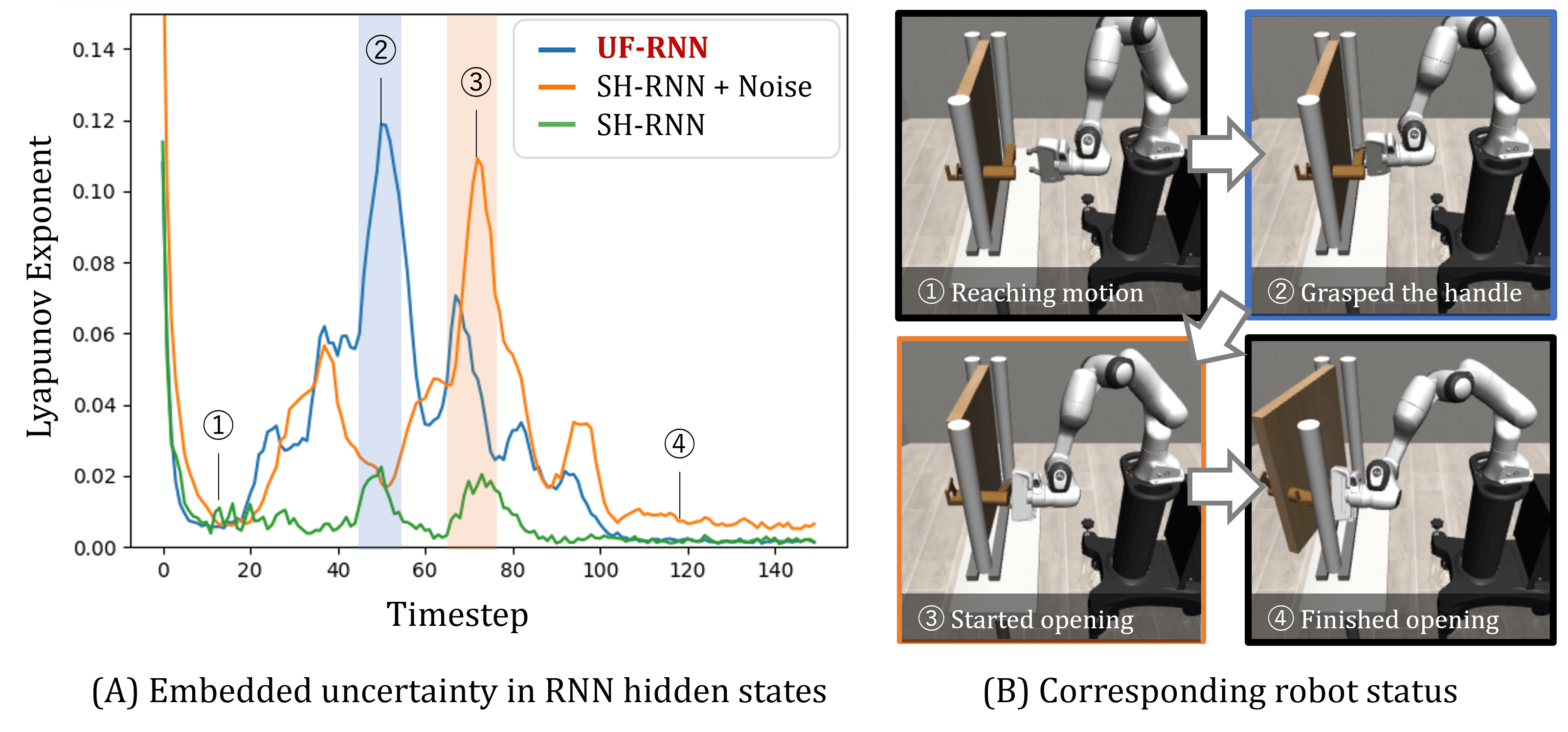}
    \caption{Transition of Lyapunov exponents compared between training model types. Highlighted area display distinct peaks of the exponent for each model, where higher values indicate that highly chaotic properties are structured at the respective timestep. The structurization of chaotic properties also display at what timing the model considered the environment to have high uncertainty.
    }
    \label{fig:sim_lyapunov}
\end{figure*}

We evaluated our proposed UF-RNN model on a door-opening task conducted in a simulated environment, as shown in Fig.~\ref{fig:sim_env}. Three different door types were included: one that must be pushed, another that must be pulled, and a third that must be slid. In each case, the doorknob must first be twisted before the door can be opened. During testing, the simulator randomly assigns one of these three door types, which are visually indistinguishable. Consequently, the robot must infer the correct door type based on real-time sensory feedback, of vision and joint angles, during the manipulation process. As the training data \textbf{only} includes successful motions for each door-opening motion, a situation without the knowledge of the door-type becomes an epistemically uncertain setup for the model.

\begin{figure*}[tbp]
    \centering
    \includegraphics[width=0.9\linewidth]{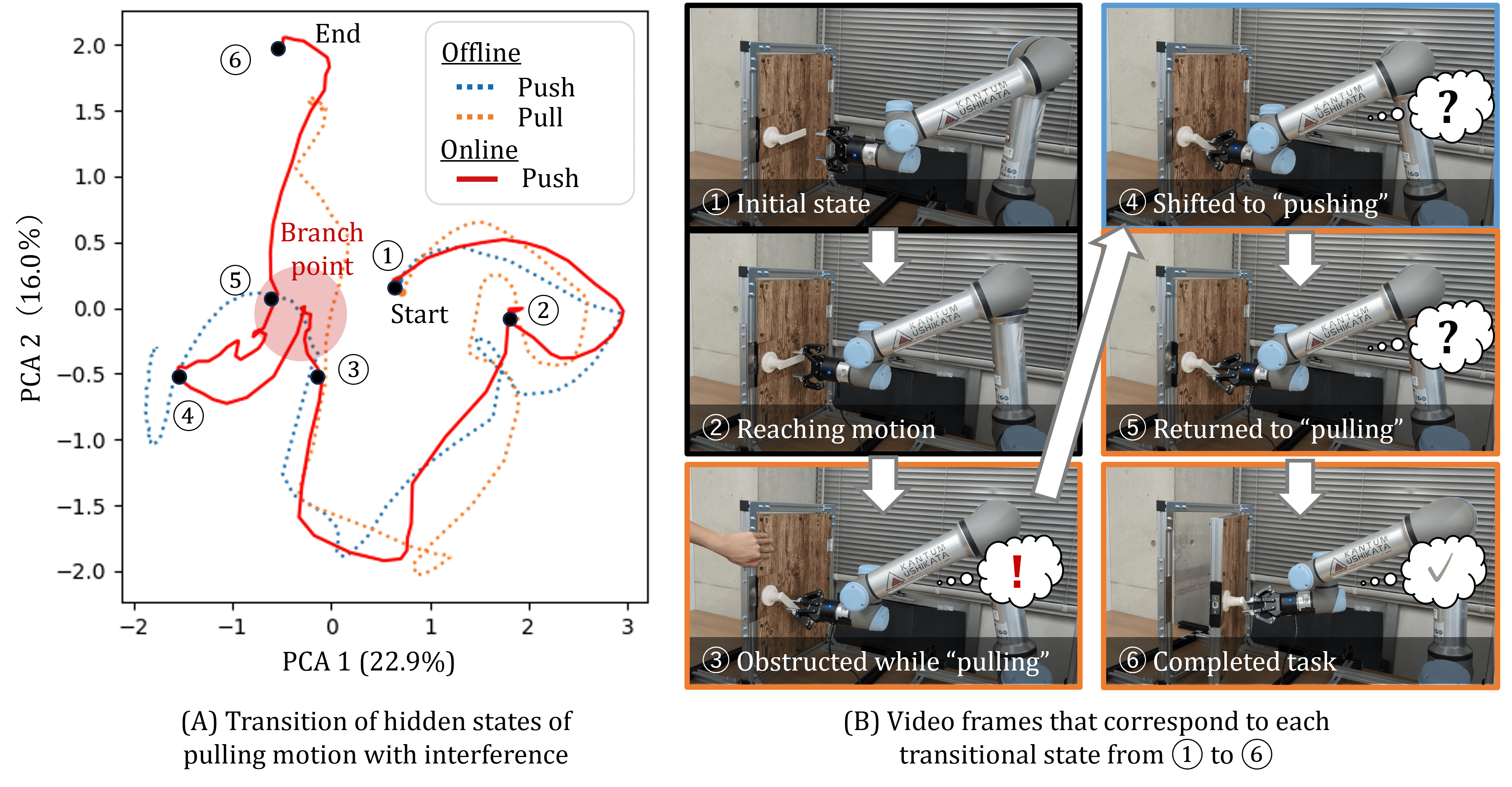}
    \caption{ The transitioning hidden states of UF-RNN on real-world experiment. The graph plots the hidden states of $LSTM_{shared}$, which is compressed to two dimensions using principal component analysis. During online prediction (solid line), the robot was interfered while attempting to open a pull-door, which caused the model to explore through its hidden space for possible solutions.
    }
    \label{fig:real_pca}
\end{figure*}

\subsection{Task Setup and Data Collection}
All training and testing scenarios were implemented in the Robosuite simulator \cite{robosuite2020}, using a 7-DoF robotic arm equipped with a gripper. We collected demonstration data by manually teleoperating the robot via a 3D mouse. Each demonstration sequence contained RGB camera images of size $128 \times 128 \times 3$ as well as joint-angle readings from the robot arm. We recorded 5 demonstration trajectories for each of the three door types, resulting in a total of 15 sequences. Each trajectory lasted 150 timesteps, recorded at 10 Hz.

\subsection{Training Procedures}
Our model was trained for 3000 epochs using the Adam optimizer with a learning rate of 0.0001 and a batch size of 5. We employed the foresight module described in Section \ref{sec:fs_module}, which predicts $N = 5$ noise patterns per timestep, and the maximum of foresight steps was set to $T=10$.

\subsection{Comparative Baselines}
We compared UF-RNN against two baselines:

\begin{enumerate}
  \item SH-RNN: A standard SH-RNN with stochastic properties from S-RNN \cite{murata2013learning} model and no foresight mechanism.
  \item SH-RNN+noise: The same SH-RNN architecture, but with random noise injected into the hidden states at each timestep, but without any foresight-based selection process.
\end{enumerate}

The second baseline serves to analyze whether simple random perturbations could replicate the exploration benefits of our foresight module.

\subsection{Success rates}
We first measured the success rate for each door type, conducting 10 trials per door type (30 trials in total). A trial was considered successful if the robot fully opened the door—i.e., achieved the correct motion sequence of knob-twisting plus push/pull/slide—to a threshold angle. Figure~\ref{fig:sim_success_rate} shows the success rates of all three models across training epochs. The UF-RNN began to learn all three motion types earlier than the baseline models, converging to approximately 80\% success after the 1300th epoch. In contrast, SH-RNN exhibited low success rates and rarely learned more than one or two types of door-opening motions reliably. SH-RNN+noise did occasionally discover all three motions, but at inconsistent epochs and with increased instability.

\subsection{Analysis of RNN Hidden States}
Figure~\ref{fig:sim_pca} illustrates the temporal evolution of the RNN hidden states for each RNN models.
The figure displays the result of Principal Component Analysis (PCA) on the hidden states over the entire trajectory of the predicted motions. The dotted lines represent offline predictions, and the solid lines correspond to the online execution. The proposed model’s hidden states structured different attractor points for each door-opening motions (c.f. offline trajectories), which during inference, was used for reproducing the robot motion. Such attractors were also used for exhibiting a branching behavior depending on the presented door type (c.f. online trajectory). At the branching point, the hidden state perturbed within a small area, producing trial-and-error motions of opening a door (push / pull / slide). Such perturbation was due to the increase of the predicted \textit{uncertainty}, which in UF-RNN, feedbacks to increase the exploratory behavior to resolve. In other words, the model temporarily explored multiple paths to different attractors before settling on the correct one, suggesting that the model effectively leveraged internal uncertainty to select the appropriate action sequence.

In contrast, both baseline models struggled to construct sufficiently distinct attractor basins within the hidden state space. The SH-RNN either collapsed onto a single suboptimal attractor or failed to show clear separations among the three door motions. The SH-RNN+noise model displayed promising branching during early training but ultimately failed to converge the hidden states onto correct attractors consistently, indicating that random perturbations alone were insufficient without the foresight-driven selection mechanism.

\subsection{Lyapunov Exponent and Chaotic Properties}
To investigate the role of chaotic dynamics in the hidden state evolution, we computed the Lyapunov exponent on the foresight prediction at each timestep. The Lyapunov exponent quantifies how small perturbations in the hidden state diverge or converge over time; higher values indicate greater sensitivity to initial conditions (i.e., stronger chaotic behavior). In this case, the intensity of the exponent indicates the possibility of various futures that the model expects, starting from each timestep.

As shown in Fig.~\ref{fig:sim_lyapunov}, the UF-RNN and SH-RNN+noise models both exhibited peaks in Lyapunov exponents at key task stages. For UF-RNN, these peaks aligned with critical branching points, such as grasping the doorknob. For SH-RNN+noise, peaks occurred during early door-opening motions but were not effectively harnessed for deciding the correct policy, owing to the lack of a foresight mechanism to guide noise-based exploration. In contrast, SH-RNN maintained a low and relatively constant Lyapunov exponent, indicating that the hidden states remained near a single attractor without diverging to explore alternative motion strategies. This is likely due to the lack of random noises, which aids the RNN hidden state to capture divergent dynamics \cite{laskey2016shiv} and structure chaotic properties \cite{matsumoto1983noise}.

Hence, while both noised models learned chaotic state transitions, only UF-RNN demonstrated “active” usage of chaos—where uncertainty was primarily directed to policy branching rather than to random fluctuations in observed sensory data. This difference reflects the foresight module’s role in coupling uncertainty with action strategies, thereby enabling the robot to explore policy alternatives selectively and converge on the most suitable door-opening approach.

\section{Real-World Experiment}

\subsection{Setup}
We further validated our proposed model on a real-world door-opening task (Fig.~\ref{fig:real_pca}). In contrast to the simulator experiment, the door could only be opened by either pushing or pulling after twisting the handle, and no sliding mechanism was involved. This is due to the difficulty in minimizing the effect of appearance changes using the slide-door. We employed a UR5e robotic arm with a gripper, resulting in a total of 7 degrees of freedom, where the gripper opening is treated as an additional DoF. An RGB camera was installed behind the robotic arm, collecting RGB images that cover the entire scene of the door and the robot.

To introduce positional variability, we horizontally shifted the door by 5~cm in both directions. This shift was intended to simulate slight misalignments in real-world installations. We collected 30 demonstration trajectories in total, each consisting of RGB images at a resolution of $128 \times 128 \times 3$ and joint angles recorded at 10\,Hz, each lasting 100 timesteps. The final dataset included 15 demonstrations for the push-door configuration and 15 for the pull-door configuration, each with slight positional offsets.

\subsection{Results}
We compared UF-RNN with the SH-RNN+noise baseline in this setting. Both models were tested at random initial positions within the trained workspace, ensuring that the door remained within the robot’s reachable zone. A trial was considered successful if the robot successfully twisted the handle and opened the door by a specified angle.

Overall, both models showed robust performance, successfully selecting the appropriate push or pull motion in most trials. Particularly, both models succeeded in performing the corresponding motions at all trials, except for few that failed to grasp the door knob owing to visual misrecognition. SH-RNN+noise also performed better here than in the simulation, likely because only two possible door-opening strategies (push vs. pull) needed to be differentiated. Nevertheless, qualitative observations suggested that UF-RNN converged more steadily during training, exhibiting fewer oscillations in the learned policies and requiring fewer epochs to achieve high success rates.

\subsection{Behavior Under External Interference}
To assess adaptive behavior in the presence of unforeseen disturbances, we conducted an interference experiment in which an operator manually held the door closed for a few seconds during a pull maneuver (Fig.~\ref{fig:real_pca}). This interference created an unexpected obstacle that prevented the robot from completing the pulling motion immediately.

We observed that UF-RNN was capable of performing adaptive behaviors in response to the obstruction. When the door was held, the robot initially displayed a “confused” pulling action but then transitioned through short push motions before returning to a pull strategy that succeeded once the door was released. From an analysis of the robot’s hidden states (via PCA of the RNN latent space in Fig.~\ref{fig:real_pca}(A)), we found that the model explored multiple attractors during the obstruction, reflecting a form of chaotic branching analogous to the simulator experiments.

In contrast, the SH-RNN+noise model tended to persist with the pulling motion, failing to diverge to an alternative course of action despite the physical blockage. While the injection of random noise created some variability in the hidden states, there was no foresight mechanism to guide the model toward exploring a push strategy. These results underscore the importance of coupling internal uncertainty with action selection, as realized by the foresight module in UF-RNN.

\subsection{Discussion}
In real-world applications, minor positional offsets, unmodeled dynamics, and external interference are common sources of uncertainty. The above results indicate that although both UF-RNN and SH-RNN+noise can handle small variations in the door properties (placement and opening directions), UF-RNN’s foresight-driven approach facilitates more consistent adaptation during unexpected disturbances. Moreover, the reduction from three door types (in simulation) to two (in reality) helped clarify that the model’s stable training is not simply a result of fewer potential actions, but rather the product of its effective exploration and policy-branching mechanism.

\section{CONCLUSION}
In this paper, we introduced UF-RNN, a novel recurrent architecture that integrates a Foresight module to proactively manage uncertainty in robotic decision-making. The Foresight module refines the hidden state by simulating multiple future trajectories and selecting the one with the lowest predicted variance, effectively coupling internal stochasticity with action selection. 

We evaluated UF-RNN on door-opening tasks in both simulated and real-world environments. Experimental results showed that, unlike conventional stochastic RNNs, our model captured chaotic properties in high-uncertainty situations and leveraged them to explore alternative actions whenever faced with ambiguous conditions (e.g., unknown door types or external interference). This explorative capacity facilitated more robust and adaptive behavior, resulting in higher success rates and smoother convergence during training.

Looking ahead, we plan to extend UF-RNN to more dynamic and complex environments that exhibit hierarchical uncertainties, such as multi-stage manipulation tasks in partially observable settings. Additionally, we aim to deepen our theoretical analysis of how foresight-driven predictions align with the free-energy principle, particularly regarding the role of chaotic attractors in biological cognition. By further integrating insights from neuroscience and variational inference, we hope to refine UF-RNN for broader applications in adaptive robotics.

\section*{ACKNOWLEDGMENT}
This work was supported by JST [Moonshot R\&D][Grant Number JPMJMS2031].

\bibliography{bibliography}
\bibliographystyle{unsrt}

\end{document}